\documentclass[10pt,twocolumn,letterpaper]{article}

 \usepackage[pagenumbers]{cvpr} 

\usepackage{float}
\usepackage{dblfloatfix}
\usepackage{booktabs}      
\usepackage{array}         
\usepackage{colortbl}      
\usepackage{xcolor}        
\usepackage{pifont}        
\usepackage{caption}       
\usepackage{graphicx}      
\usepackage{dblfloatfix}
\usepackage{cuted}
\usepackage{capt-of}
\usepackage{verbatim}
\usepackage[dvipsnames]{xcolor} 

\newcommand{\mypar}[1]{\vspace{1mm}\noindent\textbf{#1}}

\definecolor{cvprblue}{rgb}{0.21,0.49,0.74}
\usepackage[pagebackref,breaklinks,colorlinks,allcolors=cvprblue]{hyperref}

\def\paperID{42767} 
\def\confName{CVPR}
\def\confYear{2026}

\title{Circuit Tracing in Vision–Language Models:\\ Understanding the Internal Mechanisms of Multimodal Thinking}

\author{
Jingcheng Yang\textsuperscript{1*} \quad
Tianhu Xiong\textsuperscript{1*} \quad
Shengyi Qian\textsuperscript{2$\dagger$} \quad
Klara Nahrstedt\textsuperscript{1} \quad
Mingyuan Wu\textsuperscript{1}\\[0.5em]
\textsuperscript{1}University of Illinois Urbana-Champaign \quad \textsuperscript{2}Independent Researcher\\
{\tt\small \{jy95, mw34, klara\}@illinois.edu}
}

\begin{document}

\twocolumn[{%
  \renewcommand\twocolumn[1][]{#1}%
  \maketitle
}]

\begingroup
\renewcommand\thefootnote{}
\footnotetext{* Jingcheng and Tianhu contributed equally.}
\footnotetext{$\dagger$ This work was conducted independently and is not related to the author's employment at Meta FAIR.}
\endgroup

\begin{abstract}
Vision–language models (VLMs) are powerful but remain opaque black boxes. We introduce the first framework for transparent circuit tracing in VLMs to systematically analyze multimodal reasoning. By utilizing transcoders, attribution graphs, and attention-based methods, we uncover how VLMs hierarchically integrate visual and semantic concepts. We reveal that distinct visual feature circuits can handle mathematical reasoning and support cross-modal associations. Validated through feature steering and circuit patching, our framework proves these circuits are causal and controllable, laying the groundwork for more explainable and reliable VLMs. Our code and models are available at \href{https://github.com/UIUC-MONET/vlm-circuit-tracing}{https://github.com/UIUC-MONET/vlm-circuit-tracing}.
\end{abstract}   

\vspace{-20px}

\section{Introduction}
\label{sec:intro}

\begin{figure}[t]
  \centering
  \includegraphics[width=\linewidth]{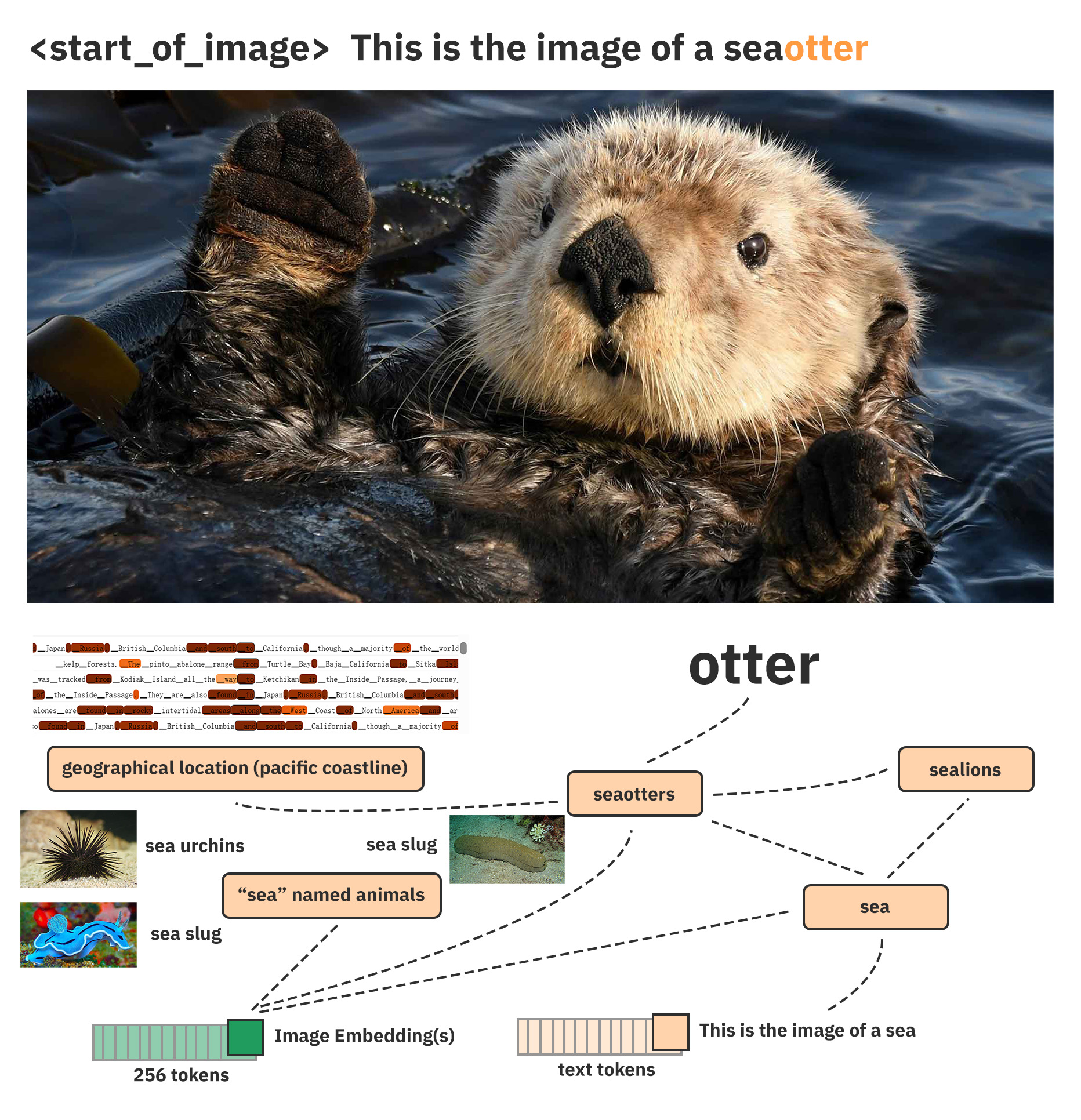}
  \caption{Given an image and a prompt, how can we extract a circuit, as an internal computation graph, of open-source vision language models such as Gemma3-4B \citep{gemma3} from Google. We introduce the first framework for successful circuit tracing in VLMs, enabling analysis of the internal circuits and association of concepts in underlying multimodal reasoning.}
  \label{fig:seaotter}
  \vspace{-20pt}
\end{figure}

The rapid advancement of vision–language models (VLMs) has fundamentally transformed how machines understand and reason about multimodal information. Contemporary VLMs such as CLIP \citep{clip}, Flamingo \citep{flamingo}, LLaVA \citep{llava}, and GPT4-o \citep{openai2024gpt4ocard} demonstrate remarkable capabilities in tasks ranging from visual question answering and image captioning to complex visual reasoning and embodied AI applications. These models seamlessly integrate visual perception with linguistic understanding, enabling machines to answer questions about images, generate detailed descriptions of visual scenes, and even perform multi-step reasoning that requires coordinating information across modalities. Despite these impressive empirical successes, a critical question remains largely unanswered: \emph{How do these models actually work internally?}

Understanding how VLMs work is essential for building trustworthy, controllable AI. Although they are used in high-stakes areas like medical imaging, autonomous driving, and content moderation, their decision-making remains opaque. This lack of interpretability makes it hard to diagnose errors, mitigate biases, and ensure alignment with human values. It also limits scientific insight—reverse-engineering their mechanisms can reveal how vision and language interact and guide the design of more capable, efficient architectures.

Recent work in interpretability has begun revealing the internal algorithms used by language models through attention visualization \citep{attention-vis}, probing, and circuit discovery \citep{circuitmethod, circuit-tracer}. Yet these methods focus almost entirely on text-only models. VLMs introduce deeper challenges: they must integrate two modalities with different statistics and semantics while discovering meaningful visual–linguistic correspondences. How VLMs bind visual features to tokens, implement cross-modal reasoning, or coordinate visual and linguistic attention remains largely unknown—posing a more complex frontier than interpretability work in single-modality text models or early visual interpretability research \citep{Visualize-CNN, olah2017feature}.

This work introduces the first framework for successful circuit tracing in VLMs, enabling systematic analysis of the internal computational mechanisms underlying multimodal reasoning. Our approach builds on recent advances in interpretability for language models. Specifically, we leverage transcoders \citep{transcoder} to decompose neural representations into interpretable features, and combine these with attribution-based circuit discovery methods \citep{circuitmethod, circuit-tracer, sae_dashboard} to map causal relationships between features. We are the first to extend these techniques to the multimodal setting. In doing so, we address the unique challenges posed by vision–language integration and develop new methods to trace information flow from visual inputs through the model's reasoning process to final outputs.

Extending circuit tracing to the multimodal domain is far more challenging than we thought, but brings richer insights and opportunities in VLMs than we anticipate.

\noindent \textbf{Make Circuit Tracing Happen in VLMs}. We build three key components in VLMs: 

\begin{enumerate}
\item \textit{Transcoders in VLM}: We insert and train per-layer-transcoders into VLMs to decompose multimodal, polysemantic representations into interpretable, monosemantic features. (\cref{sec:transcoders}) 

\item \textit{Attribution Graph}:  We trace attribution graphs that incorporate image-embedding residuals to reveal causal relations between features in the resulting multimodal computational graph. (\cref{sec:attribution}) 

\item \textit{Circuit Discovery of Visual and Text Tokens}: We propose attention analysis to interpret unnamed multimodal features, allowing for fully interpretable multimodal circuit discovery.
(\cref{sec:interpretation}) 
\end{enumerate}


\noindent \textbf{Representative Insights from Multimodal Circuits}. These circuits are not mere post-hoc correlations but represent genuine causal mechanisms which help interpret VLM internal computation. In Section \ref{sec:findings} we include:
\begin{enumerate}
\item \textit{Hierarchical Integration of
Visual and Semantic Concepts}: Features that simultaneously encode both semantic and visual concepts emerge only in higher layers of the network.
\item \textit{Highly Interpretable Case Studies}: for Visual Math reasoning, Six Finger Hallucination, Association between Mars and Space Shuttle.
\item \textit{Distinct Visual Latent Space in Language Model}: We find that visually similar features cluster and co-activate, indicating that the language-model component of the VLM preserves a distinctly visual representation space.
\end{enumerate}

\noindent \textbf{Intervention with Multimodal Circuits}. Intervention means changing the activation of features in the circuits, enabling more opportunities:
\begin{enumerate}
\item \textit{Steering}: We modify the activation of certain features to observe the impact it has on the final output. (\cref{sec:steering}) 

\item \textit{Circuit Patching}:  We transfer entire patches of one circuit to another circuit of similar structure and function, to see whether the transplanted circuit performs the same.
(\cref{sec:interventionexperiments}) 
\end{enumerate}

\noindent In summary, our contributions are threefold: we establish the first circuit tracing framework for VLMs, uncover key insights into multimodal reasoning with circuits, and demonstrate through intervention experiments the potential for circuit-based model manipulation and control.

\section{Related Work}

\mypar{Mechanistic interpretability in LLMs} aims to causally reverse-engineer how model components implement behavior, going beyond earlier correlational interpretability methods such as probing and attention analysis \citep{tenney2019bert,clark2019bert,jain2019attention}. In language models, this literature includes circuit-level analyses of specific behaviors (e.g., induction heads, IOI, and factual recall/localization) \citep{elhage2021transformercircuits,olsson2022induction,wang2022ioi,meng2022rome,meng2022memit}, intervention-based localization methods such as activation patching and path patching \citep{zhang2023activationpatching,goldowskydill2023pathpatching}, and feature-level approaches motivated by superposition, where sparse autoencoders are used to recover more interpretable features than individual neurons \citep{geva2021kv,elhage2022superposition,bricken2023monosemanticity}. The field has also become more scalable through partially automated circuit discovery (ACDC) and shared tooling platforms such as Neuronpedia \citep{conmy2023acdc,neuronpedia}.


\mypar{Sparse autoencoders and transcoders.} Neural representations are often polysemantic, responding to multiple unrelated concepts \citep{scherlis2025polysemanticitycapacityneuralnetworks}. Sparse autoencoders (SAEs) \citep{sparsify2025} mitigate this by learning sparse, overcomplete decompositions that yield more interpretable, often monosemantic features \citep{sae}. Transcoders \citep{transcoder} extend this idea by replacing MLP layers with SAEs trained to match the MLP’s outputs, preserving the model’s computation while exposing feature-level structure. This enables end-to-end attribution and circuit tracing. While transcoders have been used to analyze language models \citep{transcoder}, their application to VLMs remains unexplored.

\mypar{Vision–language model interpretability.} While interpretability methods have advanced significantly for language models, understanding vision–language models remains challenging. Most existing work on VLM interpretability focuses on high-level analysis: attention visualization between image and text tokens \citep{attention-vis}, probing for learned visual concepts \citep{olah2017feature}, and analyzing cross-modal alignment \citep{clip}. More recently, several works have attempted to localize visual knowledge in VLMs \citep{nikankin2025taskdifferentcircuitsdisentangling} or understand how visual information is processed through the network \citep{cauldron}. However, these methods typically analyze individual components in isolation or rely on correlation-based analysis rather than causal circuit discovery. Our work is the first to apply circuit tracing methodology to VLMs, enabling systematic identification of the computational mechanisms underlying multimodal reasoning.
\section{Circuit Tracing in VLMs}
\label{sec:method}

In this section, we present our framework for circuit tracing in vision--language models. We first describe how we train transcoders to decompose MLP activations into sparse, interpretable features (Section~\ref{sec:transcoders}). We then explain how attribution graphs are computed to capture causal relationships between features across layers (Section~\ref{sec:attribution}). Next, we detail our approach to analyzing and interpreting learned features through activation patterns and attention visualization (Section~\ref{sec:interpretation}). We describe our circuit discovery algorithm that identifies minimal computational subgraphs responsible for specific capabilities (Section~\ref{sec:discovery}). Finally, we present intervention and steering methods to causally validate discovered circuits (Section~\ref{sec:steering}).

\begin{figure*}[t]
    \centering
    \includegraphics[width=1\linewidth]{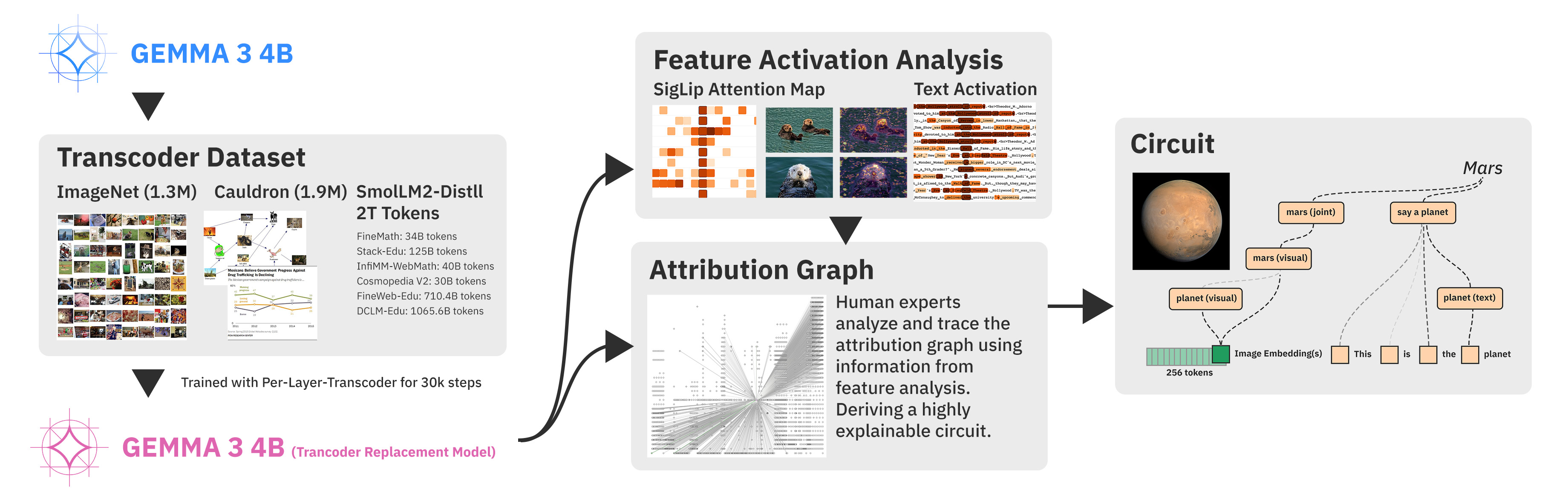}
    \caption{A summary of our method, we first train transcoders for Gemma-3-4b-it on our curated dataset, yielding a replacement model with monosemantic features. Then, we utilize feature activation analysis to obtain information that would provide interpretable information about the features. Then, we generate an attribution graph of a given prompt, and utilize human experts to derive the final circuit.}
    \vspace{-1.5em}
    \label{fig:pipeline}
\end{figure*}

\subsection{Transcoders}
\label{sec:transcoders}

Sparse Autoencoders (SAEs) \citep{sae} provide a sparse, interpretable basis for transformer activations, but they do not cleanly expose causal relations between features, limiting circuit discovery. Whereas SAEs are trained directly to reconstruct a transformer's activations, transcoders \citep{transcoder} replace a transformer's MLP sublayer with a sparse autoencoder, maintaining computational equivalence while enabling feature-level analysis.

For each MLP layer in the vision--language model, we train a transcoder consisting of an encoder and decoder. The encoder maps the MLP input $x \in \mathbb{R}^{d_{\mathrm{model}}}$ to $d_{\mathrm{feat}} \gg d_{\mathrm{model}}$ learned latent features via:
\begin{equation}
z(x) = \mathrm{ReLU}(W_{\mathrm{enc}} x + b_{\mathrm{enc}}),
\end{equation}
\noindent where $W_{\mathrm{enc}} \in \mathbb{R}^{d_{\mathrm{feat}} \times d_{\mathrm{model}}}$ and $b_{\mathrm{enc}} \in \mathbb{R}^{d_{\mathrm{feat}}}$ are learned parameters. The decoder reconstructs an approximation to the original MLP output:
\begin{equation}
\mathrm{TC}(x) = W_{\mathrm{dec}} z(x) + b_{\mathrm{dec}},
\end{equation}
\noindent where $W_{\mathrm{dec}} \in \mathbb{R}^{d_{\mathrm{model}} \times d_{\mathrm{feat}}}$ and $b_{\mathrm{dec}} \in \mathbb{R}^{d_{\mathrm{model}}}$.

Unlike the original transcoder implementation \cite{transcoder}, which uses an $\ell_1$ penalty to encourage sparsity, we follow \citet{sparsify2025} and enforce sparsity directly via $\mathrm{TopK}(z(x), k)$, retaining only the $k$ largest activations. This removes the need for a sparsity coefficient and yields more stable training with consistent sparse features.
We evaluate reconstruction quality using the \emph{Fraction of Variance Unexplained} (FVU):
\begin{equation}
\mathrm{FVU}
= 
\frac{\tfrac{1}{n}\sum_{i=1}^n (y_i - \hat{y}_i)^2}
     {\tfrac{1}{n}\sum_{i=1}^n (y_i - \bar{y})^2}
= \frac{\mathrm{MSE}}{\mathrm{Var}(y)}.
\end{equation}
Where $y$ is $\mathrm{MLP}(x)$ (the original MLP output) and $\hat{y}$ is $\mathrm{TC}(x)$. Training minimizes only the reconstruction error, while sparsity is controlled solely by the choice of $k$ (rather than through a tunable penalty). Each transcoder feature is defined by a paired encoder column and decoder row, contributing additively to the output. The expanded latent space ($d_{\mathrm{feat}} = N_{\mathrm{latents}} \cdot d_{\mathrm{model}} \cdot N_{\mathrm{layers}}$)
enables polysemantic MLP representations to be factorized into sparse, approximately monosemantic, and thus interpretable features.

While transcoders have been extended to model cross-layer structure \citep{circuitmethod}, we use the per-layer formulation as originally introduced by \citet{transcoder}. After training a transcoder for each MLP block, we construct a replacement model by substituting each MLP with its corresponding transcoder, yielding a network expressed entirely in sparse, learned latent features. Because transcoders only approximate the original MLPs, we explicitly track the reconstruction residual
\begin{equation}
e(x) = \mathrm{MLP}(x) - \mathrm{TC}(x),
\end{equation}

\noindent computed from cached MLP outputs and included as a separate error node in the circuit graph. This accounts for approximation error without altering the forward pass of the replacement model.

\begin{figure}[t]
  \centering
  \includegraphics[width=\linewidth]{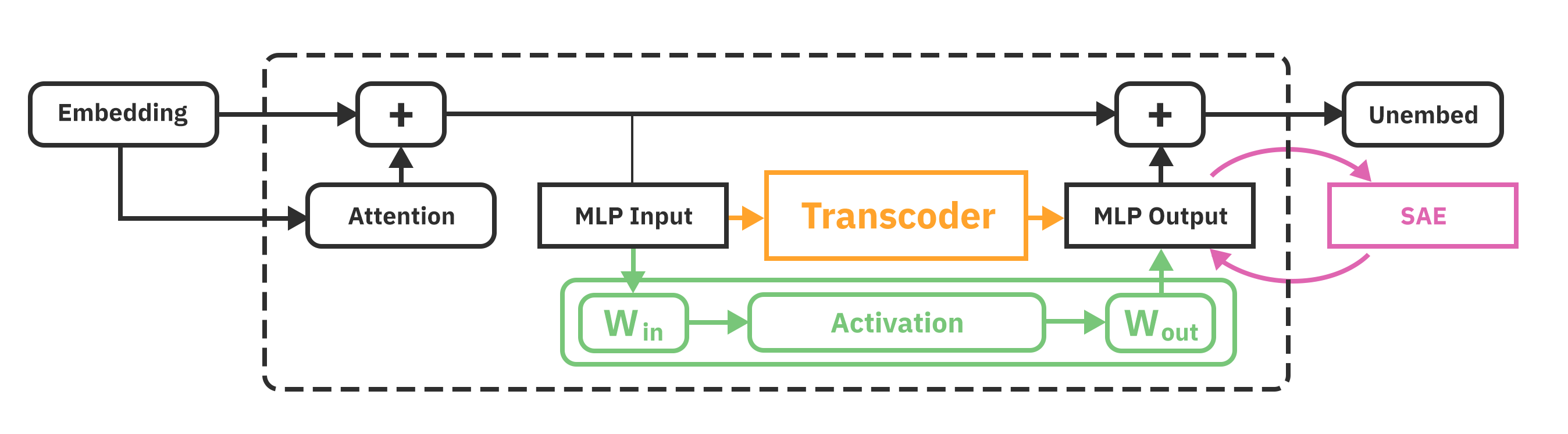}
  \vspace{-2em}
  \caption{Transcoder vs SAE; SAEs learn to reconstruct model activations, whereas transcoders imitate MLP sublayers’ input-output behavior.}
  \vspace{-1.0em}
  \label{fig:transcoder}
\end{figure}

\subsection{Attribution Graphs}
\label{sec:attribution}

To trace circuits, we compute an attribution graph as introduced by \citet{aibiology} and \citet{circuitmethod}, adapted for per-layer transcoders by \citet{circuit-tracer}, that linearly decomposes how features contribute to activations in upper layers and, ultimately, to the output logits on a fixed prompt. Because each transcoder replaces the MLP with a sparse linear readout from features, and all nonlinearities (ReLUs, attention patterns, and normalization factors) are frozen at their values on the given prompt, the model becomes locally linear around that input.

Each node in the graph corresponds to a token embedding, an active transcoder feature at a particular (layer, position) pair, or an output logit. For a source node $s$ and target node $t$, the attribution is defined as:

\vspace{-10px}

\begin{equation}
A_{s \to t} = a_s \, w_{s \to t},
\end{equation}

\vspace{0px}

\noindent where $a_s$ is the activation magnitude of source feature $s$ and $w_{s \to t}$ is the \emph{virtual weight}—the local derivative of $t$'s pre-activation with respect to $s$'s activation. For per-layer transcoders, this weight factors into the decoder of $s$, the frozen transformer Jacobian, and the encoder of $t$:
\begin{equation}
w_{s \to t} = f_{\mathrm{dec}}^{(s)\,\top} \, J^{\blacktriangledown}_{(s)\to(t)} \, f_{\mathrm{enc}}^{(t)},
\end{equation}
\noindent where $f_{\mathrm{dec}}^{(s)} \in \mathbb{R}^{d_{\mathrm{model}}}$ is the decoder vector for feature $s$, $f_{\mathrm{enc}}^{(t)} \in \mathbb{R}^{d_{\mathrm{model}}}$ is the encoder vector for feature $t$, and $J^{\blacktriangledown}_{(s)\to(t)} \in \mathbb{R}^{d_{\mathrm{model}} \times d_{\mathrm{model}}}$ is the residual-stream Jacobian of the original transformer with stop-gradients on all nonlinearities (LayerNorm, attention softmax, and ReLU).

Since the model is linearized, the pre-activation of each node is exactly the sum of its incoming attributions:

\begin{equation}
h_t = \sum_{s \in \mathrm{pred}(t)} A_{s \to t},
\end{equation}

\noindent yielding a complete additive explanation of how earlier representations influence later ones. We construct the full directed attribution graph $G = (V, E)$ for each prompt, where nodes $V$ include all token embeddings, active features, and output logits, and edges $E$ have weights $A_{s \to t}$. We prune edges with negligible attribution ($|A_{s \to t}| < \epsilon$) to produce a sparse, interpretable graph.

\subsection{Feature Interpretation and Attention Analysis}
\label{sec:interpretation}

\textbf{Feature activation analysis.} To understand what each transcoder feature represents, we analyze its activation patterns across a diverse dataset of vision--language inputs. For each feature $f_i$, we collect the top-$k$ activating examples—image--text pairs $(I, T)$ that produce the highest activation $z_i$—and examine commonalities. We compute activation statistics including the feature's activation frequency (fraction of inputs where $z_i > 0$), mean activation magnitude, and position distribution (which tokens or image patches most frequently activate the feature).


\mypar{Vision encoder attention maps.}
While feature activations on text tokens are directly interpretable, activations on image tokens remain opaque. To visualize which image regions the SigLIP vision encoder attends to, we compute attention-rollout maps for each of the visual embeddings passed to the language model.

In the SigLIP vision-encoder for Gemma 3, it processes a resized \(896 \times 896\) image and reduces the representation to a fixed set of \(N_v = 256\) visual tokens that are then input to the language model.  

We compute attention-rollout maps over the last \(K\) self-attention layers of the vision tower. Let the encoder output \(T\) tokens in total (including possibly a class token), of which \(N_v\) are the visual patch tokens entering the language model. For each of the last \(K\) layers \(\ell\), we select the fraction \(q\) of attention heads with lowest mean entropy (i.e., the most focused heads), average their \((T\times T)\) attention matrices to obtain \(\bar A^{(\ell)}\), add an identity residual, and row-normalize to form a row-stochastic matrix \(\tilde A^{(\ell)}\). The rollout is  
\[
R \;=\; \tilde A^{(L-K+1)} \,\tilde A^{(L-K+2)} \cdots \tilde A^{(L)} \;\in\; \mathbb R^{T \times T}.
\]
We then extract the \(N_v \times N_v\) sub-matrix \(R_{\mathrm{vis}}\) corresponding to the visual tokens and interpret each row as an attention distribution over the 256 visual token positions. To provide spatial localization, we recover the original spatial grid of image patches (e.g., \(g_h \times g_w\)) from the encoder, pool tokens within non-overlapping blocks of size \(b \times b\), reshape into a \(g_h \times g_w\) grid, upsample to the input image resolution, and normalize each map. The resulting grayscale heat-maps indicate which image regions are most attended to in the vision tower. We chose not to add additional refinements, denoising or post-processing so as to present the raw information for human experts to analyze.

\subsection{Circuit Discovery}
\label{sec:discovery}

A \emph{circuit} is an abstract representation of the computational graph that explains a model's output
logits for a given input. Instead of tracking every feature activation, through explaining the feature representations as described in Section~\ref{sec:interpretation}, we can group features of similar functions into shared nodes, resulting in a simplified graph that explains the model's behavior.

This mechanistic structure highlights the key computational pathways and interactions that drive the model’s behavior, allowing us to steer and intervene in the model's activations towards predictable directions.

While there are now methods for large-scale, partially automated circuit discovery \citep{conmy2023automated},
a human-annotated circuit remains the most accurate and interpretable representation of a model’s
mechanism. In this work we therefore use human experts to discover and annotate circuits.

\subsection{Intervention and Steering}
\label{sec:steering}

To study how individual transcoder features influence model behavior, we directly modify feature activations during the forward pass and observe the resulting changes in the model’s output.

\mypar{Feature representation.}
At layer~$\ell$ and position~$t$, the transcoder produces feature activations
\[
z_{\ell,t,i} \in \mathbb{R},
\]
and each feature has an associated decoder vector $d_{\ell,i} \in \mathbb{R}^{d_{\mathrm{model}}}$ that determines how changes to that feature are written back into the residual stream.

\mypar{Interventions.}
An intervention specifies a target value $v_{\ell,t,i}$ for a feature. During the forward pass we compute\vspace{-5px}
\[
\Delta z_{\ell,t,i} = v_{\ell,t,i} - z_{\ell,t,i}(x),
\]
and apply the corresponding residual update
\vspace{-5px}
\[
h_{\ell,t} \;\gets\; h_{\ell,t} + \Delta z_{\ell,t,i}\, d_{\ell,i}.
\]
This simply adjusts the model’s internal representation as if the feature had taken the desired value.

\mypar{Circuit patching} refers to directly overwriting selected internal activations during the forward pass, or transplanting entire sub-circuits from another circuit, and observing how the model’s output changes. In our case, we patch transcoder features by setting them to chosen target values (e.g., zero for ablation or a positive constant for amplification) at specific layers and positions. We observe whether transferring a patch from Circuit A onto B would recreate similar behaviors of A for B.

All experiments are performed on the transcoder-replaced model, and we qualitatively inspect the differences in predictions to understand the behavioral role of the intervened features.
\section{Experiments}
\label{sec:experiments}

\mypar{Model.} We conduct our circuit tracing experiments on Gemma-3-4B-it~\cite{gemma3}, a state-of-the-art open-source vision–language model. Gemma-3-4B-it processes images using a SigLIP~\cite{zhai2023siglip} vision encoder, which produces patch tokens that are concatenated with text tokens and fed into a transformer decoder. This architecture enables joint visual–linguistic processing and makes the model suitable for multimodal reasoning.


\mypar{Architecture.}
The model consists of a SigLIP encoder with patch size $14$, applied to $896{\times}896$ input images.
This produces a grid of $64 \times 64$ patches, yielding $4096$ initial patch tokens \citep{gemma3}. These tokens are
then pooled into $256$ soft image tokens before being appended to the text sequence for decoding.
Gemma-3 models employ bidirectional attention over image tokens, allowing the model to attend to the
full image context throughout processing.

The decoder is a 34-layer transformer with $d_{\mathrm{model}} = 2560$, $34$ attention heads, and
MLP dimension $d_{\mathrm{ff}} = 10240$.

\subsection{Training Transcoders}
\label{sec:exp_setup}

\begin{table}[h!]
\centering
\setlength{\tabcolsep}{9.5pt}
{\fontsize{8pt}{13pt}\selectfont
\resizebox{\linewidth}{!}{%
\begin{tabular}{c|c|c}
\toprule[1pt]
Name & Content & Size \\
\midrule
SmoLIM2-135M-10B~\citep{smolLM} & 2048 text tokens & 144{,}000 \\
ImageNet~\citep{imagenet} & 1 image; Caption & 144{,}000 \\
Cauldron~\citep{cauldron} & 1 image; QA Text & 72{,}000 \\
\bottomrule[1.2pt]
\end{tabular}%
}}%
\vspace{-5px}
\caption{Datasets used for Transcoder Training.}
\label{tab:transcoder_data}
\vspace{-5px}
\end{table}

\mypar{Datasets.} We train transcoders on a diverse corpus of vision--language examples to ensure broad coverage of multimodal concepts and reasoning patterns. Our training dataset consists of a base set of pure text dataset for broad coverage of various domains \citep{smolLM}, alongside ImageNet \citep{imagenet} for rich visual features and Cauldron \citep{cauldron} consisting of QA and visual reasoning tasks. For the cauldron split, we sampled from its 50 subsets evenly to ensure even coverage.


\mypar{Implementation details.} 
For transcoder training, we implemented our framework with support for Gemma-3 on top of Sparsify \citep{sparsify2025}.

\mypar{Training configurations.} 
For every MLP layers in Gemma-3-4B-it, we train a separate transcoder with $d_{\mathrm{feat}} = [N_{\mathrm{latents}} \cdot d_{\mathrm{model}} \cdot 34]$ features. $N_{\mathrm{latents}}$ is the number of features in the hidden layer of a single transcoder block, which we then multiply by 34, the number of decoder layers in Gemma-3-4B. During the forward pass of the transformers, we only activate the top $k$ latents as described in Section~\ref{sec:transcoders}, for our experiment we set $k$ to 48.

\begin{figure}[t]
  \centering
  \includegraphics[width=\linewidth]{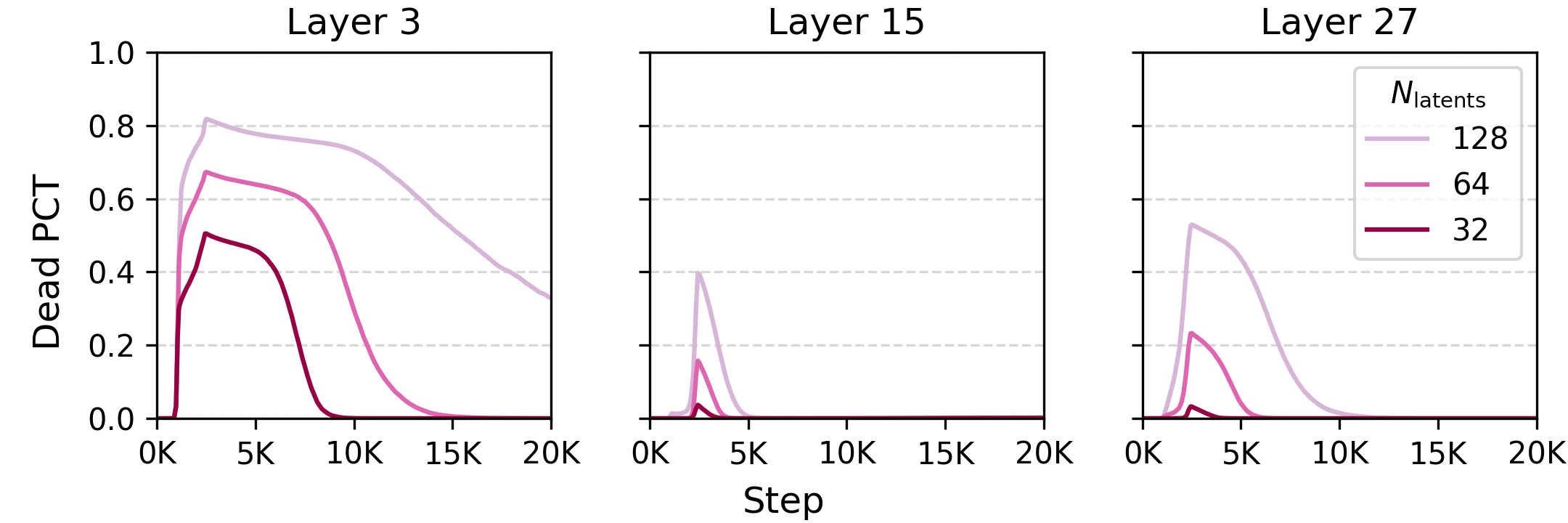}
  \includegraphics[width=\linewidth]{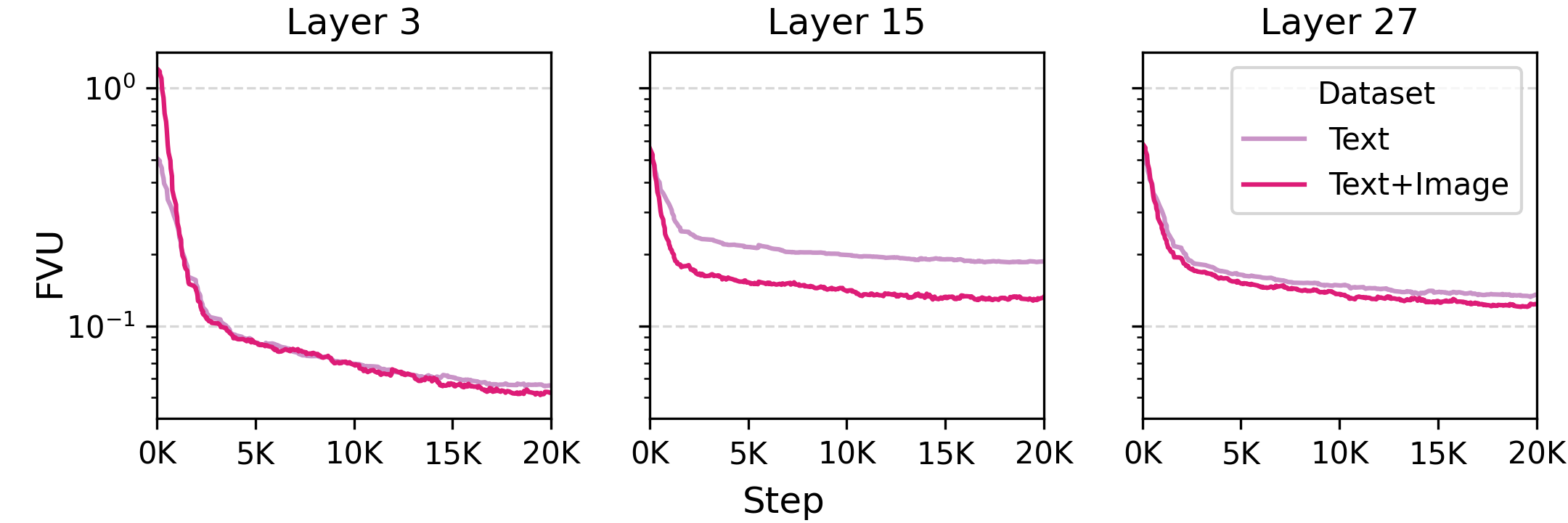}
  \vspace{-2em}
  \caption{\textbf{Top:} Percentage of dead latents (Dead PCT) across layers for different values of $N_{\mathrm{latents}}$. \textbf{Bottom:} Fraction of variance unexplained (FVU) when training transcoders on a text-only dataset compared to our multimodal split.}

  \vspace{-2.0em}
  \label{fig:ablation}
\end{figure}


Transcoder training uses the AdamW \citep{adamw} optimizer with a learning rate
$
 {2 \times 10^{-4}} \times {\sqrt{\frac{2^{14}}{N_\mathrm{latents} \times d_\mathrm{model}}}}
$,
with $N_{\mathrm{latents}}$ = 64, $d_\mathrm{model}$ = 2560. We also apply a linear warmup over the first
$N_{\mathrm{warmup}} = 1{,}000$ steps. Training uses a batch size of 12 and runs for 30{,}000 steps
on 8~H100 GPUs for approximately 60 hours.

\mypar{Expansion Factor.}  We trained transcoders with different expansion factor $N_{\mathrm{latents}} \in \{32, 64, 128\}$. As shown in the top panel of Figure~\ref{fig:ablation}, the choice of $N_{\mathrm{latents}}$ substantially affects the proportion of dead latents. We define a latent as \emph{dead} if it fails to activate above a minimal threshold for an extended portion of training. A high fraction of dead latents indicates that the model is not utilizing the full capacity of the latent space, suggesting limited diversity or overlap of concepts represented within the corresponding features. The curves also illustrate how this behavior varies across layers: early layers (e.g., Layer 3) exhibit substantially higher dead-latent ratios, whereas mid-network layers (e.g., Layer 15) show far denser activation patterns.

\mypar{FVU and multimodality.} The bottom panel of Figure~\ref{fig:ablation} compares the FVU obtained when training transcoders on our text-only split (SmoLIM2-135M-10B) versus our full multimodal (text+image) dataset. Although FVU is computed in-domain and does not directly reflect out-of-domain explainability, transcoders trained with multimodal supervision consistently achieve lower FVU across all layers. This suggests that visual features provide additional constraints that make the underlying representations more explainable. The gap is largest in the middle layers (e.g. Layer~15), consistent with our broader findings (\cref{sec:findings}) that this is where visual information is integrated into unified semantic representations; thus, explaining these circuits requires both modalities. At higher layers, the representation space becomes more uniform, and the FVU gap correspondingly decreases.


\subsection{Computing Attribution Graphs}

\mypar{Implementation details.} We computed attribution graphs using the circuit tracer library by \citet{circuit-tracer}, where we extended support for Gemma-3 VLMs. Images in Gemma-3 are not masked \citep{gemma3}, achieving full attention and allowing the model to see every part of the image in a bidirectional manner. Our attribution graphs are adapted to reflect this. To properly monitor residual streams and activation of individual transformer blocks, we utilize the TransformerLens library by \citet{nanda2022transformerlens}, with extended support for VLMs \citep{nikankin2025taskdifferentcircuitsdisentangling}.


\mypar{Attribution.} For attribution graph computation, we continuously add nodes and edges to our attribution graph until it reaches a cumulative influence thresholds $0.8$ and $0.98$ respectively.

Circuit discovery uses the top-$m$ features ranked by partial influence, where $m$ controls the maximum number of feature nodes ($m = 7500$), and we select up to 10 logit nodes whose cumulative probability mass is at least $0.95$. All attribution experiments run in bfloat16 precision to reduce memory usage. Computing an attribution graph for a simple image-text QA task takes 20 minutes on a single H100 GPU.

\begin{figure}[t]
  \centering
  \includegraphics[width=\linewidth]{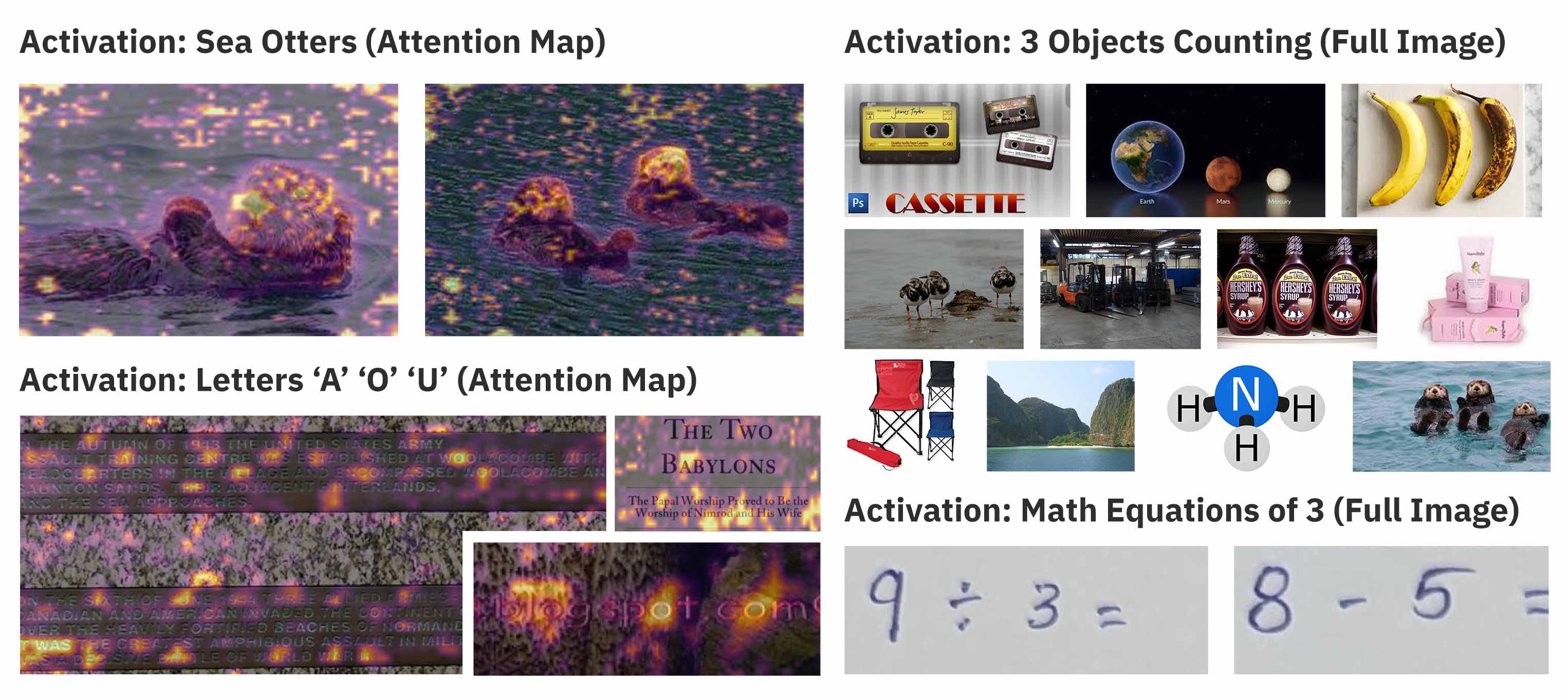}
  \caption{Feature activations on images with a mix of feature annotations resulting from full images (right) and feature analysis resulting from noisy attention maps (left).}
  \label{fig:attention}
  \vspace{-10px}
\end{figure}

\subsection{Feature Analysis}
\mypar{Datasets.} Our dataset for feature analysis is a subset of our transcoder training dataset. Including image--text pairs sampled from ImageNet (18,000 images) and Cauldron (10,000 images)~\cite{imagenet,cauldron} as well as a text-only split of 10,000 $\times$ 2048 tokens \citep{smolLM}. For circuit discovery and validation, we use a task-specific dataset consisting of a small, curated sample of 20-100 internet images that are semantically related to the image we want to trace. Our results demonstrate this to be a highly effective and efficient method at improving a graph's interpretability.

\mypar{Implementation Details.} For computing activations, we base our method upon SAE Dashboard \citep{sae_dashboard} for LLM feature activation analysis. We insert hookpoints at every MLP sublayer's input and output to record activation patterns using TransformersLens extended with VLM support \citep{nanda2022transformerlens, nikankin2025taskdifferentcircuitsdisentangling}. For activations upon image tokens, we store the index of the image tokens and the image in order to retrieve its attention map during circuit discovery.




\mypar{Ad Hoc Feature Analysis.} Whereas circuit tracing tools \citep{circuit-tracer,neuronpedia} for LLMs precompute all feature activations, we selectively compute only the features in a given attribution graph, which usually contains under 2000 features given our pruning parameters, significantly reducing the compute and storage costs of caching activation data. 

While our curated dataset allows for general interpretability of various features in the attribution graph, it may not be sufficient. In our attribution graph analyzing the image of a sea otter, further calculating the feature activations on a curated, small dataset of 30 sea otter images significantly increased the feature's interpretability and allows us to be more certain of a feature's function.

Computing feature activation on a single attribution graph with approx. 1000 features takes 20 H100 GPU hours on our dataset, with a negligible increase when injected with a curated dataset.

\mypar{Attention Maps.}  We extract the custom SigLip encoder from Gemma-3-4B-it, and use that to compute attention maps for images using our method described in Section~\ref{sec:interpretation}. Computing attention maps is fast, and we pre-compute attention maps for all the 28,000 images, which takes about 2 hours on a single H100 GPU and results in approx. 2TB of single channel masks that are recombined with the base image when queried.

\begin{figure}[t]
  \centering
  \includegraphics[width=\linewidth]{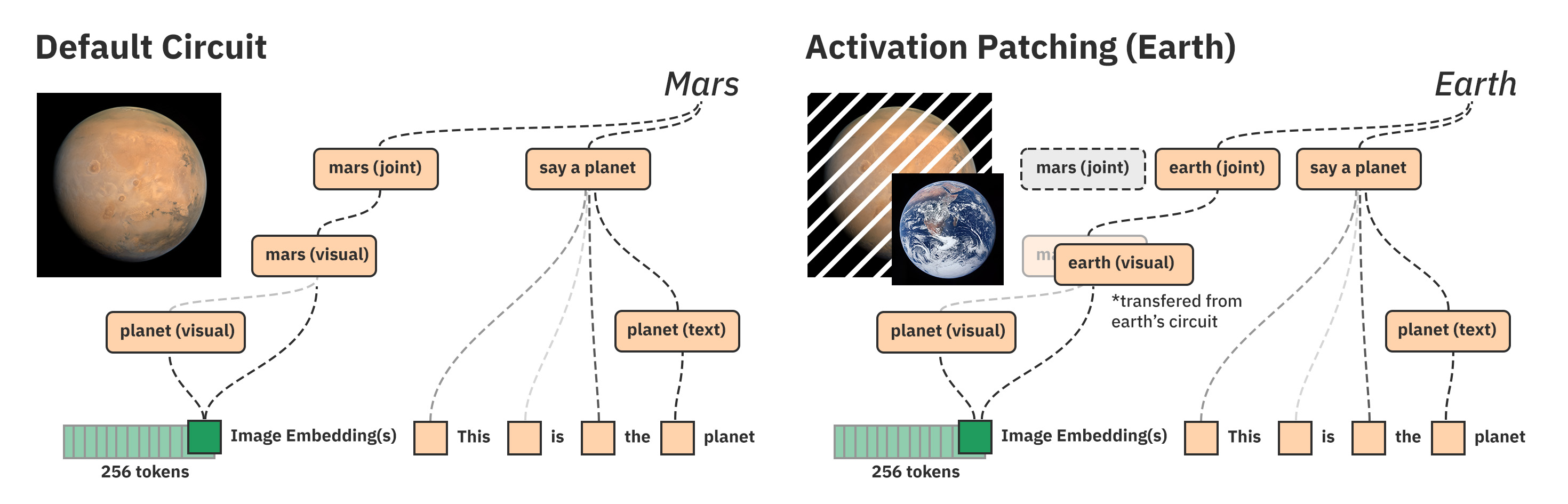}
  \vspace{-2em}
  \caption{We conduct circuit tracing on the same prompt \textit{This is the planet} with images of mars and earth. We then conduct activation patching by suppressing the mid-layer visual features of mars in the mars circuit, and set the activation of earth visual features discovered in the earth circuit in the mars circuit}

  \vspace{-2.0em}
  \label{fig:intervention}
\end{figure}

\subsection{Finding Circuits}
To obtain the most accurate circuits for validation, we use human experts (\cref{sec:discovery}) to compute the subgraph. We manually identify features that exhibit a similar function (e.g. represent similar concepts, has similar effects on model behavior), and group them into distinct nodes. The attribution between nodes is the sum of the attribution exerted by the features in each node. The resulting graph of nodes is generally an explainable, simplified circuit.

\subsection{Intervention Experiments}
\label{sec:interventionexperiments}

We use a combination of steering and activation patching (\cref{sec:steering}) to validate our circuits. Detailed experiments are available in the appendix. One example of activation patching is shown in figure \ref{fig:intervention}, where we suppressed features representing the visual concept of Mars, and instead activated features representing the visual concept of Earth identified in the earth circuit, and the subsequent feature activations and final output all changed to earth-related concepts.
\section{Empirical Findings}
\label{sec:findings}
Our circuit tracing framework reveals several core principles underlying how VLMs such as Gemma3 integrate and manipulate multimodal information. The results validate our methodology and offer a clearer picture of the computational structure supporting vision-language reasoning.

\mypar{Hierarchical Formation of Multimodal Representations.}
We observe a progressive integration of visual and semantic information as network depth increases. Only in higher layers (emerging around Layer~20) do features jointly encode both visual and semantic concepts. Earlier layers remain largely modality-specific, supporting the progressive binding hypothesis in which cross-modal associations are gradually assembled across depth.

\mypar{Granularity and Monosemanticity Across Layers.}
Feature representations become increasingly abstract with depth. Early layers exhibit highly localized, fine-grained visual patterns—down to digits or textures—while later layers form object- and concept-level features, paralleling trends in vision models \citep{Visualize-CNN, nikankin2025taskdifferentcircuitsdisentangling} but now aligned with semantics. 

\begin{figure}[t]
  \centering
  \includegraphics[width=\linewidth]{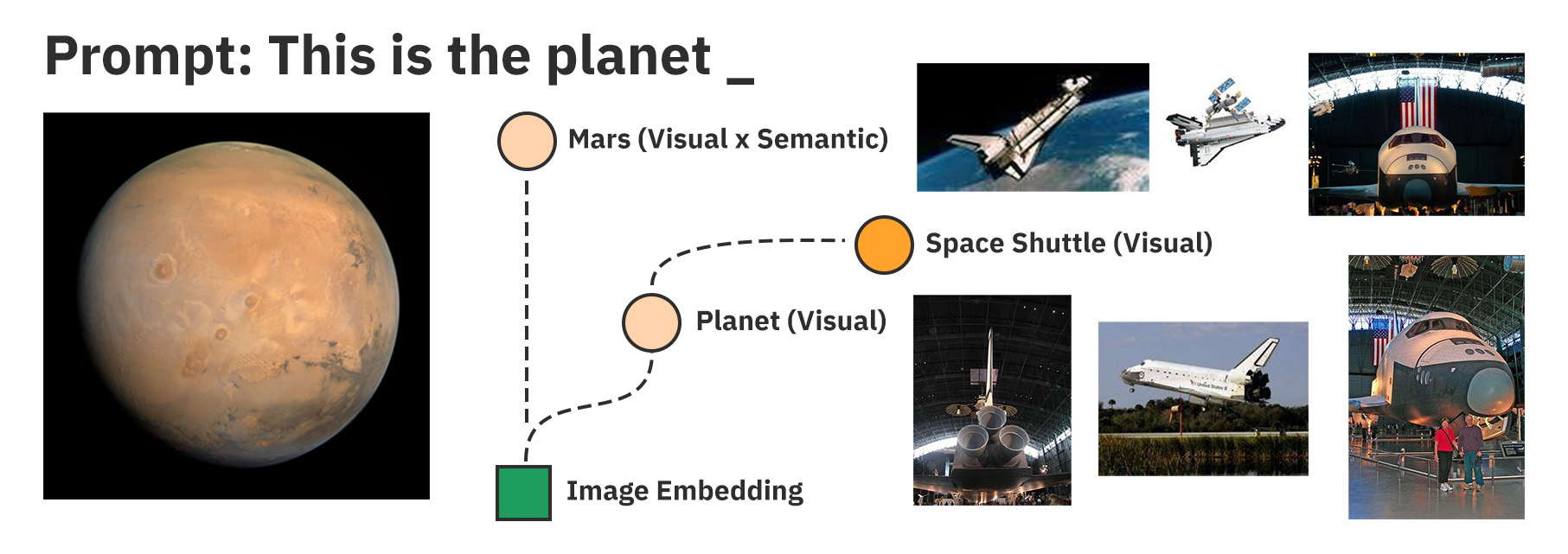}
  \vspace{-2em}
  \caption{Tracing a purely visual attribution path from an image of Mars reveals internal visual associations (e.g., “space shuttle’’) even without supporting cues. }
  \vspace{-10px}
  \label{fig:association}
\end{figure}

\mypar{Visual Circuits in Simple Mathematical Reasoning.}
For image-based arithmetic (e.g., $1+2$ rendered visually), the model appears to compute partially within visual space. Intermediate layers contain visual features corresponding to the resulting numeral (e.g., “3’’), activating across contexts. We also identify visual encodings of digit ranges and modular arithmetic patterns, echoing textual findings from \citet{aibiology}. These results suggest that simple arithmetic over images can rely on visual circuits rather than purely semantic computation.

\mypar{Understanding Hallucination: The Six-Finger Problem.}
Our tracing suggests that hallucinations—such as the six-finger case—arise from an interaction between perceptual bias and internal circuit dynamics rather than from a single failure mode. The vision encoder appears to produce embeddings that heavily emphasize generic ``hand’’ semantics, while the model’s internal circuits further amplify these features. As a result, visual features corresponding to the digit “6’’ are suppressed toward the level of unrelated numbers, whereas hand-related features strongly activate the “five’’ circuit. Although the model possesses circuits capable of visual counting (\cref{fig:attention}), these can be overshadowed by more dominant semantic and perceptual signals, showing how both encoder-level weighting and feature competition jointly contribute to hallucination.

\mypar{Parallel Visual and Semantic Pathways with Late Convergence.}
Gemma3 maintains distinct visual and semantic representation streams deep into the network. We identify visually grounded associative features—such as “space shuttle’’ activations triggered by an image of Mars (\cref{fig:association}) —reflecting internal visual associations independent of semantics. High-level layers also preserve visual similarity (e.g., consistent activations for sea otters, seals, and beavers) even when semantic categories diverge. These streams ultimately merge in the final layers, where they align into a unified multimodal representation supporting coherent reasoning.

\section{Conclusion}
This work introduces the first circuit-tracing framework for vision–language models, revealing the mechanisms underlying multimodal reasoning. Using transcoders to extract interpretable features and attribution graphs to trace causal structure, we isolate sparse circuits for tasks such as object recognition, counting, QA, and captioning. Intervention experiments confirm these circuits are causally meaningful, enabling targeted impairment and controllable steering. Beyond advancing scientific understanding, this framework provides practical tools for debugging, mitigating failures, and guiding more interpretable VLM design—supporting the development of transparent, controllable, and aligned AI systems.

\section{Limitations and Future Work.}
Despite these contributions, many important limitations remain. First, the vision-encoder attention maps used for interpreting visual features can be difficult to read: they sometimes fail to localize relevant regions or provide meaningful contextual cues, limiting their utility for feature annotation. In addition, for tasks requiring additional processing of visual inputs (e.g., math operations), it is difficult to distinguish between features that mediate the computation and features that represent its output.

Second, our use of per-layer transcoders cannot capture cross-layer superposition \citep{circuitmethod}, which could be a more significant drawback for VLMs due to the high feature density of image embeddings; we frequently observe many near-duplicate visual features firing in attribution graphs, suggesting the need for cross-layer or more adaptive transcoder designs. In addition, our work does not investigate in detail regarding specific choices of the transcoder configuration, under the assumption that they are largely the same. Research into specific SAE methods (e.g. JumpRelu, BatchTopK, etc) and varying configurations for optimal transcoder training for VLMs could prove to be useful.

Thirdly, by largely mirroring the circuit-tracing method for LLMs \citep{circuitmethod}, the computational cost for understanding and explaining multimodal features have risen significantly. While we accommodated for this limitation by computing feature activations ad-hoc, we believe methods for comprehensive or automated interpretation of features could greatly reduce the complexity of our current process. Current methods for automatic feature interpretation \citep{paulo2025automaticallyinterpretingmillionsfeatures} are computationally too expensive, and we call for future works to improve this process.

In addition, our current analysis is limited to one specific model. We have reason to suspect some complexities in circuit-tracing might be complicated by Gemma3's SigLip and bidirectional-attention mechanism. Extending this work to accommodate a wider collection of VLMs could greatly solidify findings and conclusions in our current work. 

Finally, circuit discovery currently requires substantial human effort, making it difficult to introduce quantitative evaluation or apply our method directly to model fine-tuning or improvement. Automating feature labeling, simplifying attribution graphs, or developing cross-layer transcoders could help scale circuit tracing to larger models and enable more systematic evaluation.

\section{Acknowledgements}
This research used the Delta advanced computing and data resource, which is supported by the National Science Foundation (award OAC-2005572) and the State of Illinois. Delta is a joint effort of the University of Illinois Urbana-Champaign and its National Center for Supercomputing Applications.
This work was supported by the National Science Foundation under awards CNS-2106592 and CCF-2217144. Any opinions, findings, and conclusions or recommendations expressed in this material are those of the authors and do not necessarily reflect the views of the National Science Foundation.

\newpage

{
    \small
    \bibliographystyle{ieeenat_fullname}
    \bibliography{main}
}

\clearpage
\setcounter{page}{1}

\maketitlesupplementary

\section{Transcoder Training}
\label{sec:supp_transcoders}

\begin{strip}
    \centering
    \includegraphics[width=1\linewidth]{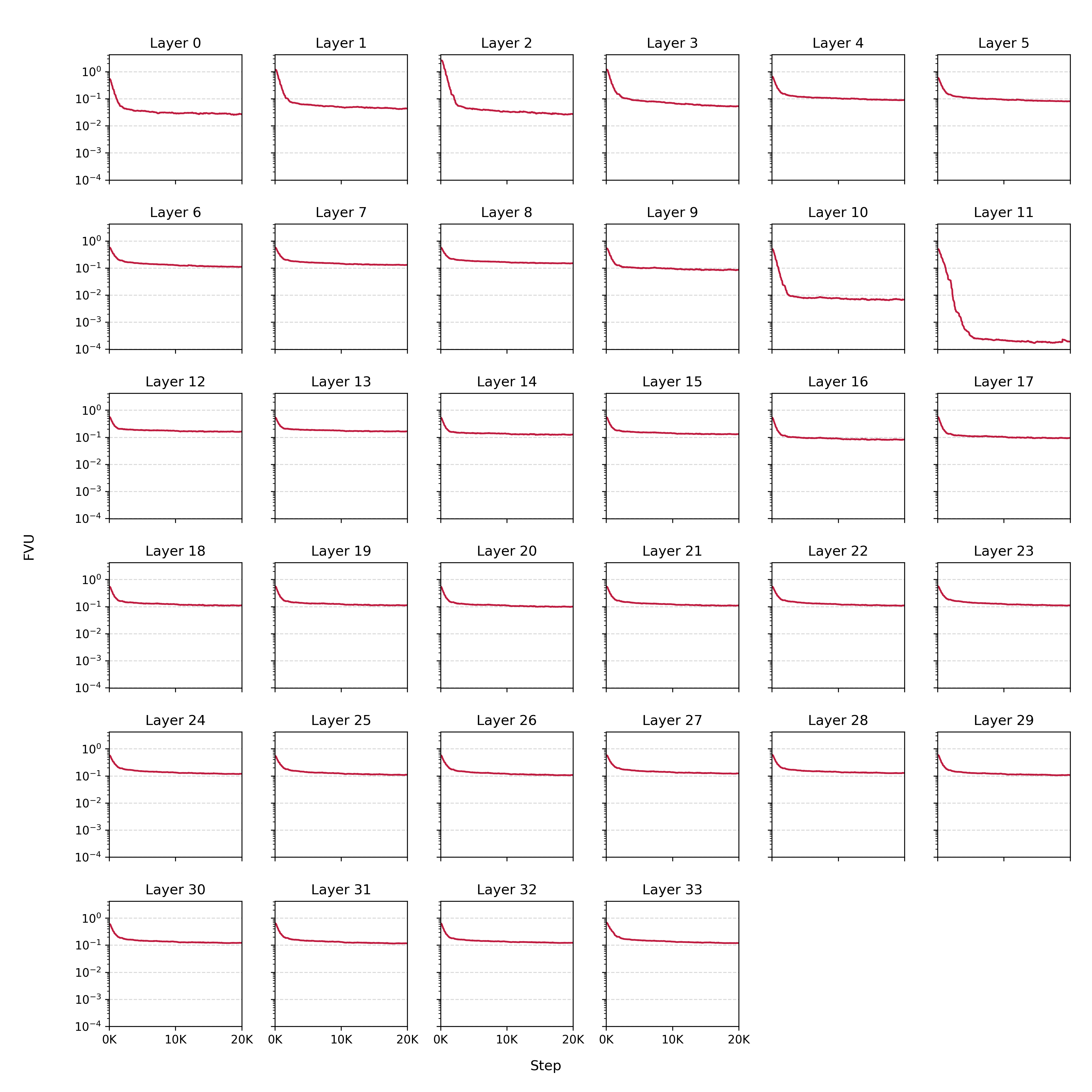}
    \captionof{figure}{FVU (Fraction of Variance Unexplained) Training curve for Gemma-3-4B-IT.}
    \label{fig:all_fvu}
\end{strip}

\begin{strip}
    \centering
    \includegraphics[width=1\linewidth]{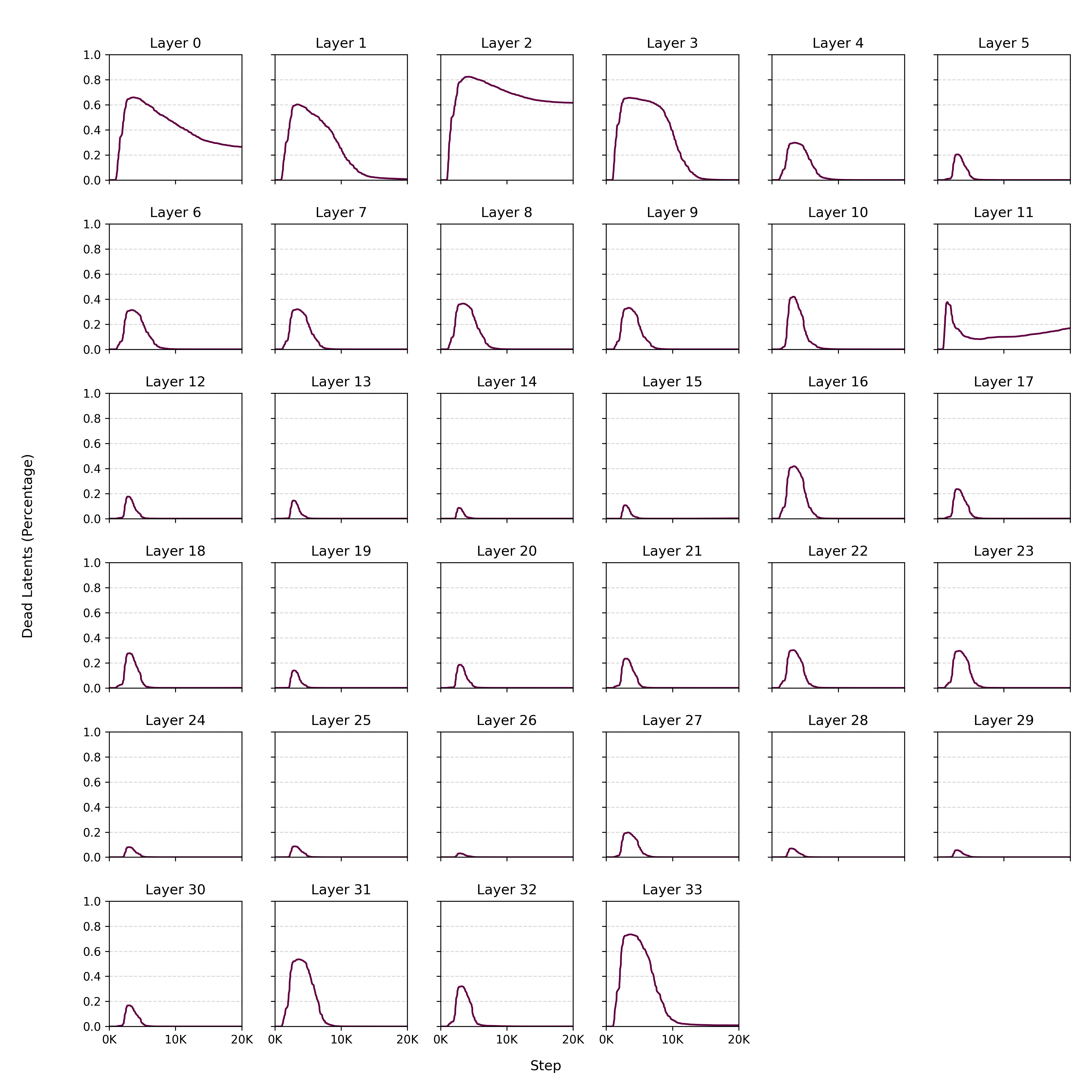}
    \captionof{figure}{Percentage of dead features over training for all layers of Gemma-3-4B-IT.}
    \label{fig:all_dead}
\end{strip}

\subsection{Dead Latents}
As shown in Figure \ref{fig:all_dead}, a feature is marked as dead if it fails to activate sufficiently over a given interval, and it is considered active once it activates sufficiently again. The percentage of dead latents provides insight into how features utilize the available expansion size. A high percentage does not necessarily imply high polysemanticity; it may instead reflect insufficient expansion capacity to support fully monosemantic features, causing the model to default to a denser cluster of max-k features that occupies less space than the nominal expansion factor.
We hypothesize that the lower layers may either represent condensed, naturally superpositioned embeddings originating from the vision encoders, or may require a much larger expansion factor to be properly disentangled. In either case, we currently lack a mature method for accurately extracting low-level representations—such as patterns, colors, or other subtle features—that we believe are present in these layers, due to the difficulty of identifying consistent patterns when examining activation example images in aggregate. We plan to follow up on this in future works.
Layer 11, being a global-attention layer, stands out from the local layers around it and creates a clear shift in the model’s training curve. Because it suddenly incorporates full-context information, its behavior differs enough to change the curvature. Its transcoder is harder to interpret for the same reason—it mixes information from across the entire sequence rather than nearby tokens. Even so, this complexity is contained within the layer, so it doesn’t disrupt the rest of the circuit-tracing process.

\section{Feature Discovery}
\label{sec:supp_feature_discovery}

Feature discovery on multi-modal inputs is significantly more expensive than on text-only LLMs. For text activations, a single token with its surrounding context is often sufficient to explain a feature’s behavior. As a result, a paragraph of a few hundred tokens can provide hundreds of useful activation examples.

In contrast, image inputs in VLMs produce far more tokens. For Gemma-3-4B, each image yields 256 tokens. Although a sufficiently dense image could, in principle, provide unique and informative signals across all 256 embeddings, most images do not contain enough complexity for all tokens to meaningfully activate distinct features. In typical or unfavorable cases, the full set of 256 image tokens may provide no more explanatory coverage than a single text token. Moreover, the vision encoder introduces additional computational overhead.

These factors are the primary reason we have not yet analyzed all VLM features. We hope to optimize our pipeline and explore alternative methods for obtaining feature activations more efficiently.

\section{Circuits}
\label{sec:supp_circuits}

\subsection{Hallucination - Fingers}

\begin{figure}[t]
  \centering
  \includegraphics[width=\linewidth]{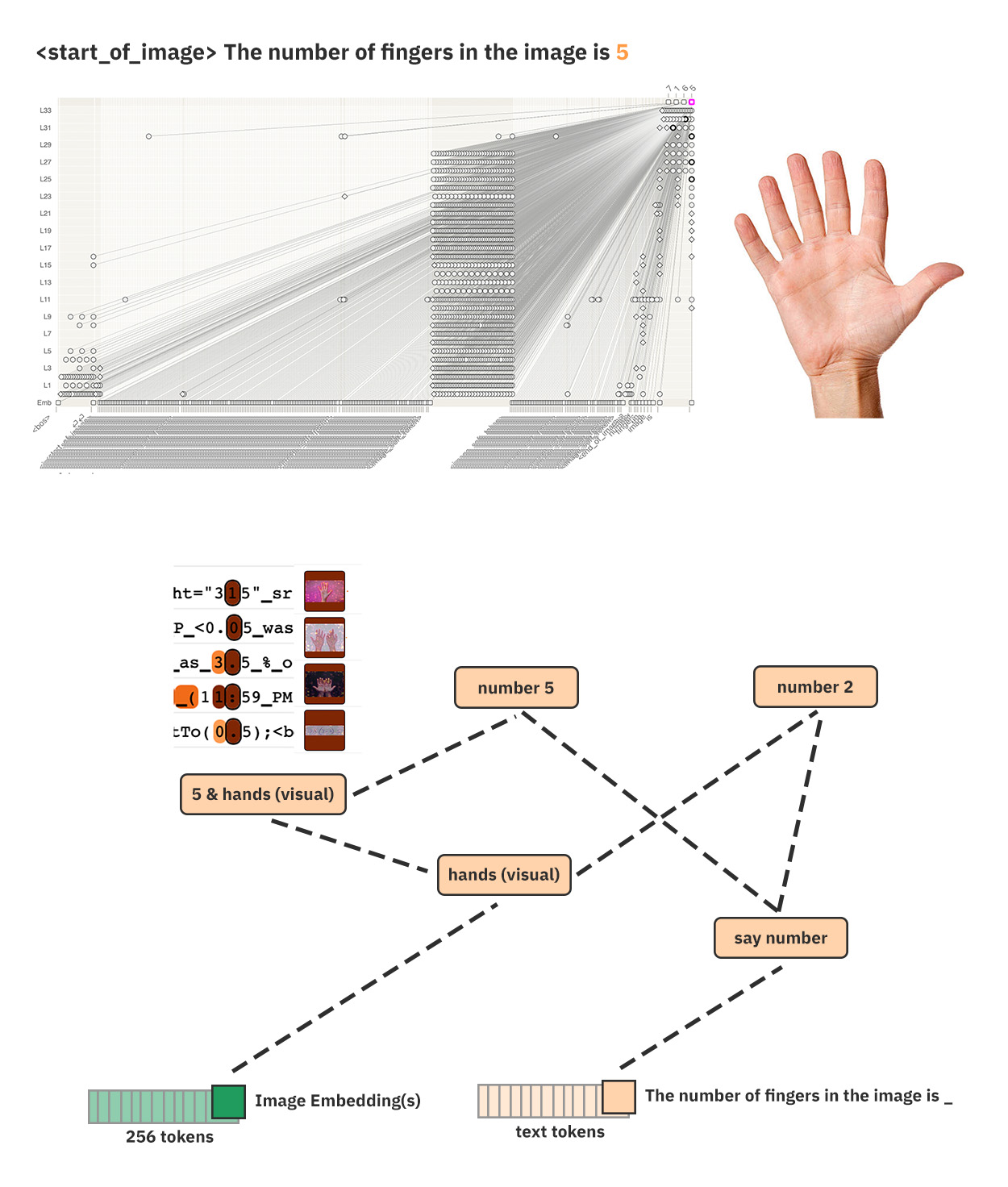}
  \vspace{-2em}
  \caption{Circuit tracing analysis of prompt "The number of fingers in the image is " with the result being 5, despite image containing six fingers. }
  \label{fig:circuit_hands}
\end{figure}

We analyzed the common hallucination in which the model predicts five fingers instead of six. Remarkably, for this model, the logit for the token 6 is no higher than for other unrelated digits (e.g., 7 or 1). Although we have previously identified features involved in visual counting tasks (e.g., features that activate on groups of three), we were unable to find any feature corresponding to the concept of six or six visual objects in the attribution graph.

Instead, we found a direct circuit between the image embeddings, features associated with the number 5, and the final output for the ‘5’ token. Notably, contrary to the typical assumption that the model would rely on a semantic “hand” concept, the circuit appears to be driven by a visual feature representing the visual concept of a hand together with the number five. In other words, in this task, the visual concept of hand activates the concept of five, rather than the model performing a robust object-counting procedure.

We also observed features related to the number 2 in upper layers, which—possibly due to the input—become activated by the concept of a hand and by features representing the function “say a number.” We hypothesize that this may arise because hands occur in pairs across animals, causing the model’s number-related circuits to be influenced by hand priors rather than meaningful counting information from the encoder. Overall, this suggests that the error may stem from the encoder failing to provide a sufficiently strong representation of a six-fingered hand—or from the model implicitly choosing to ignore such evidence.

\subsection{Caption - Sea Otter}

\begin{figure}[t]
  \centering
  \includegraphics[width=\linewidth]{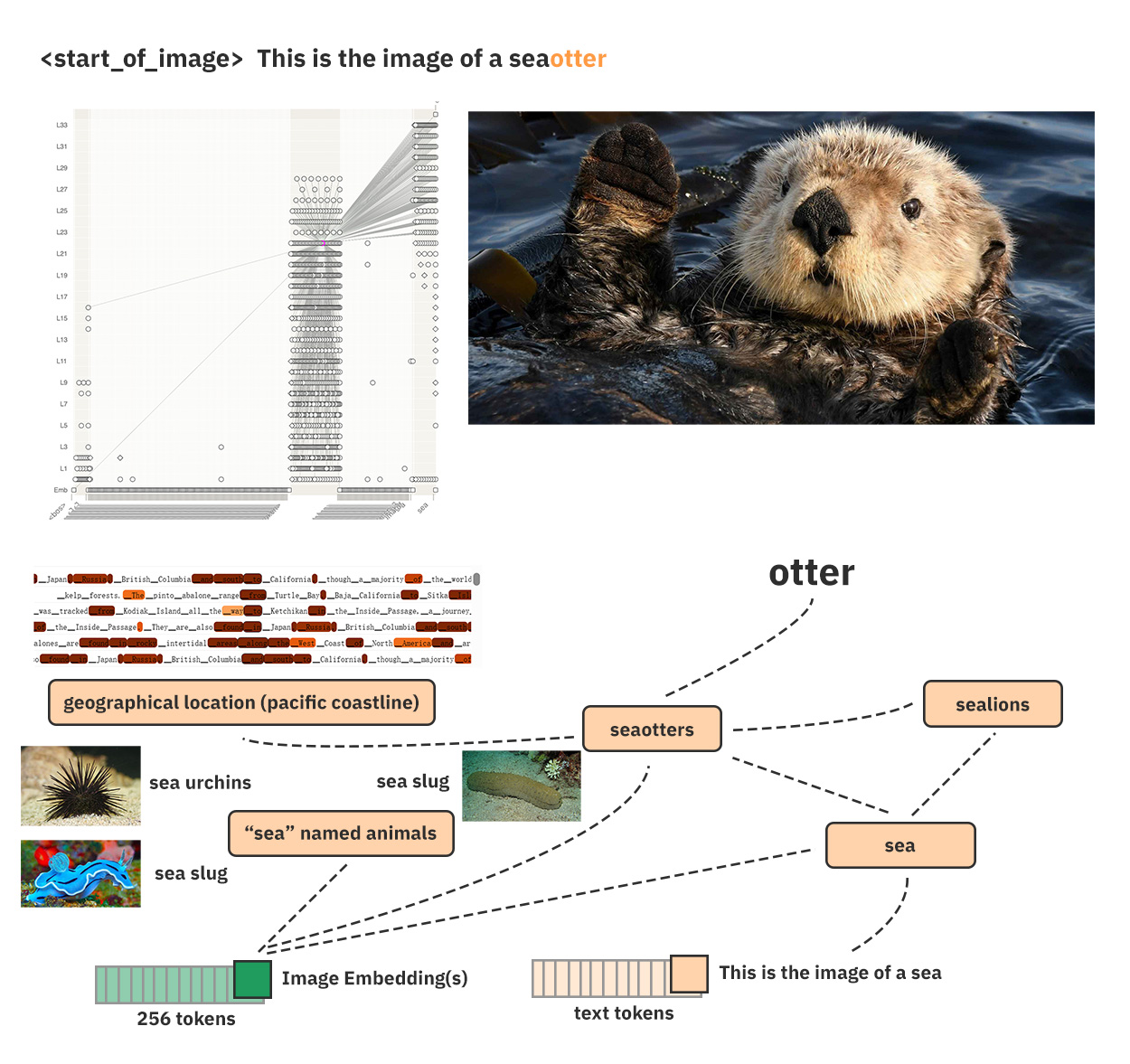}
  \caption{Circuit tracing analysis of a seaotter, with interesting feature attributions. }
  \label{fig:circuit_seaotter}
\end{figure}

We conducted circuit tracing on an image of a seaotter, which required two-step circuit analysis due to seaotter being two tokens. We observe for the token "sea" the logic was in the 90+\% range, and we analyzed the circuit prompted seaotter.

Interestingly, we also found that this image strongly activated a feature that represented sea lions - likely due to their visual similarities (the feature was also present in the first circuit without the sea token input), this shows that there may exist a purely visual latent space in the model.

We also found another feature that included visual images of animals that contained the word "sea", such as sea urchins, sea slug, and sea cucumbers. There also exists a knowledge feature of a geographical range, which just happens to be the pacific coastline from California to Japan - where sea otters primarily live in.

\subsection{Caption - Mars}

\begin{figure}[t]
  \centering
  \includegraphics[width=\linewidth]{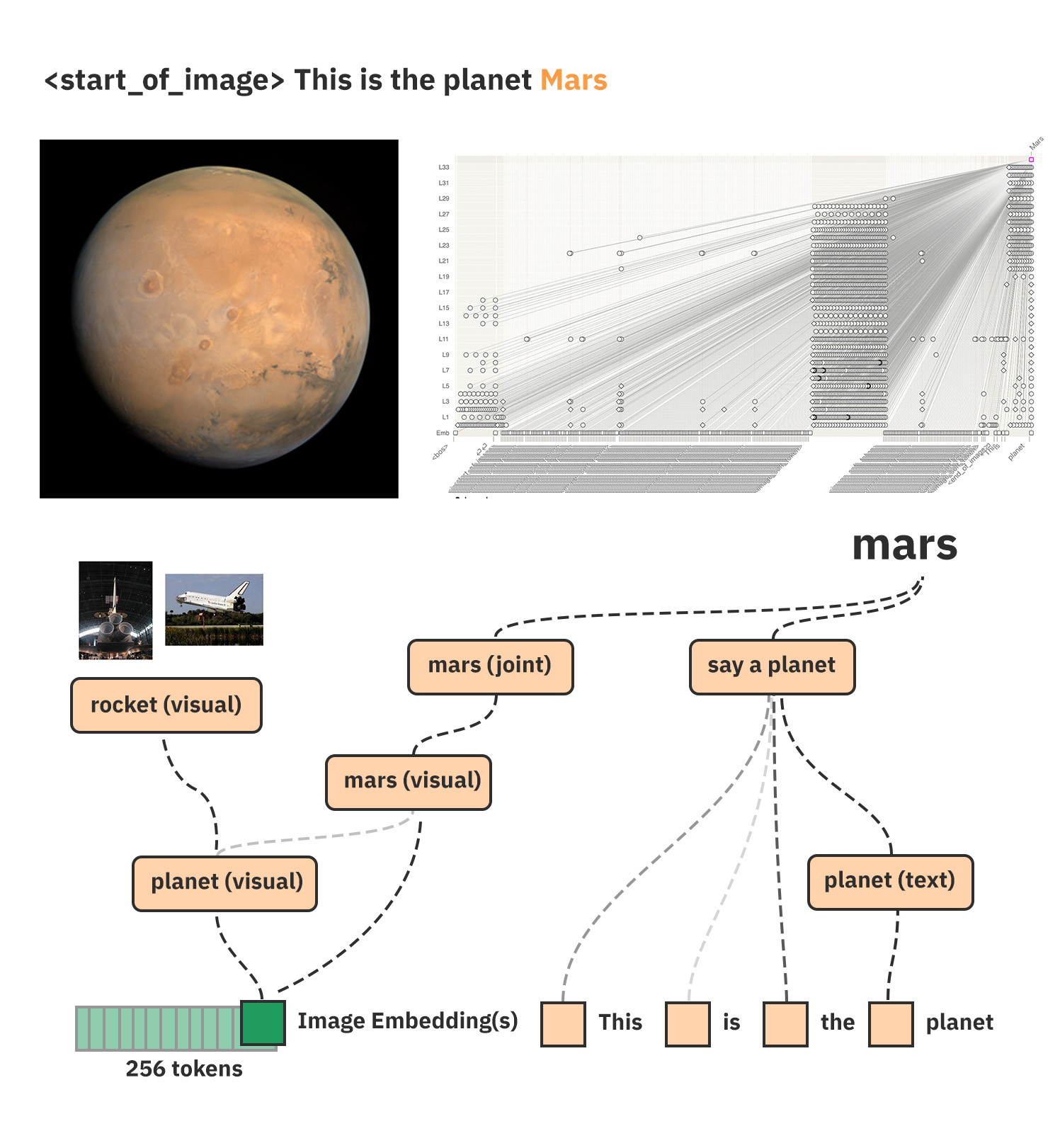}
  \vspace{-2em}
  \caption{Analysis of the mars circuit, revealing a transition from purely visual features to a semantically grounded feature that controls the final output.  }
  \label{fig:circuit_mars}
\end{figure}

We analyzed the mars prompt in Figure \ref{fig:circuit_mars}, which reveals a clean and interpretable circuit showing how raw visual representations from the encoder are gradually transformed into a unified multimodal feature representing the concept of Mars. This joint feature blends both the semantic textual notion of the word “mars” and the visual appearance of the planet, demonstrating how the model builds a shared representation that supports cross-modal grounding. The circuit further illustrates how early visual features—such as a generic “planet” detector—feed into progressively more specific features until they converge on a single joint concept that dominates the logit contribution for the final output token.

Notably, the circuit also highlights a broader associative structure present in mid-lower layers. Features representing planets reliably activate features associated with rockets and space shuttles, even though these objects do not appear in the input image. This suggests that the model has learned a latent web of visual associations that mirrors human conceptual priors: planets evoke spacecraft, just as certain animals evoke particular habitats or behaviors. These associations arise purely from visual features rather than explicit textual grounding, indicating that VLMs develop an internal “association of ideas”—where visually related objects co-activate—even before higher-level semantic features take over. This phenomenon provides insight into how VLMs integrate visual context, prior knowledge, and semantic structure when generating grounded descriptions.

\subsection{Reasoning - Simple Addition}

\begin{figure}[t]
  \centering
  \includegraphics[width=\linewidth]{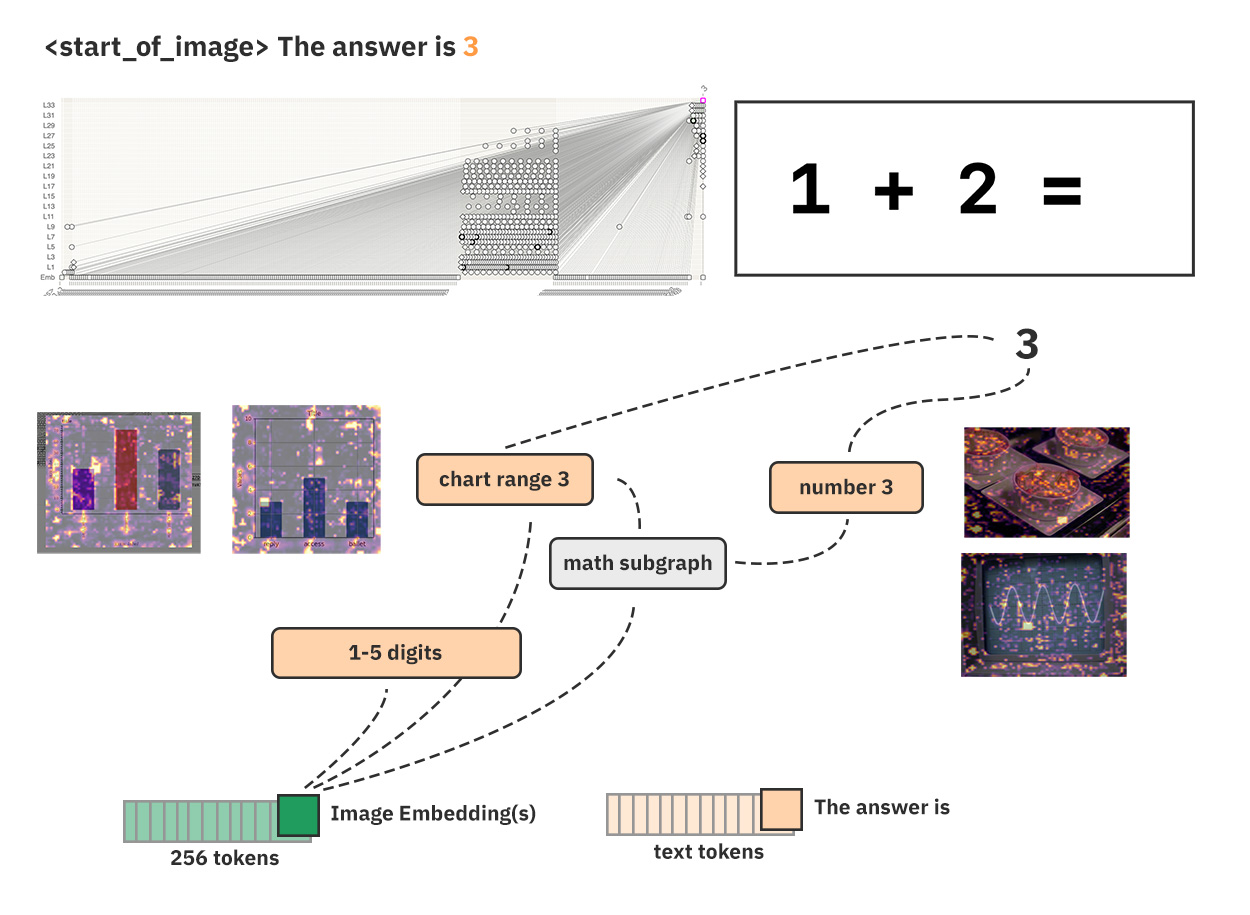}
  \caption{Circuit tracing analysis of a simple addition task. }
  \label{fig:circuit_addition}
\end{figure}

As shown in Figure \ref{fig:circuit_addition}, we analyzed a simple addition problem with an image of an incomplete equation. We found that at lower layers, the image embeddings activated features related to numerals, such as a feature representing a number between 0-5, showing that visual-semantic convergence may also occur at lower layers for low-level representations.

We also note that the model contains a very diverse feature representation of numbers. For example, we identified a feature that primarily activates on images of charts, with the attention roughly focusing on the 3 range of the axis. This feature is potentially used for visual-numerical reasoning. We also found features activating on objects in groups of three, this shows that object count information is passed from the vision encoder, and that the VLM decoder contains features that can properly represent this. We note that this is entirely absent in the finger hallucination example which prompts us to incline towards the vision encoder not encoding such information properly.

In the middle-upper layers, we discovered a group of semantic features that activates on mathematical operations, and we hypothesize that this is the subgraph that performs this operation in semantic space. Analyzing the purely semantic addition circuit is beyond the scope of our analysis, and we recommend referring to the circuit tracer paper by \citet{aibiology} for a deeper analysis.

\section{Conclusion}

In this appendix, we presented additional analyses generated using our proposed VLM circuit-tracing methodology, illustrating its ability to reveal structure across visual, semantic, and multi-modal representations. Through examinations of transcoder behavior, feature activation patterns, and several representative circuits, we showed how VLMs blend visual features with language-level abstractions—sometimes in unintended ways. These case studies highlight several recurring themes: the limitations of current vision encoders in providing robust object-level information; the presence of latent spaces that mix visual similarity, linguistic priors, and associative knowledge; and the emergence of compact, semantically meaningful features even within early or visually dominated layers. Our circuits serves as a useful tool to debug hallucinations, false knowledge, and internal mechanisms of VLMs for future improvements.

Overall, these findings demonstrate both the promise and challenges of interpreting VLMs. While our method successfully uncovers many of the mechanisms driving model behavior, it also reveals gaps—such as missing low-level disentangled features or overreliance on semantic priors—that motivate further refinement of both architectures and interpretability tools. We hope that the insights provided here, along with the methodology introduced in the main paper, will help pave the way for a more systematic understanding of multimodal reasoning, its failure modes, and the underlying circuits that support these abilities in modern VLMs.

\end{document}